\documentclass[10pt,a4paper,logo]{paper}

\usepackage[all]{hypcap}

\usepackage[authoryear, round]{natbib}

\usepackage{algorithm}
\usepackage{algpseudocode}
\usepackage{booktabs}
\usepackage{caption}               
\usepackage{subcaption}
\usepackage{graphicx}
\usepackage{threeparttable}        
\usepackage{wrapfig}               
\usepackage{multirow}
\usepackage{xcolor}
\usepackage{bbding}                
\usepackage{lipsum}                
\usepackage{amsmath}               
\usepackage{graphicx}              
\usepackage{accents}               
\usepackage{multirow}              
\usepackage{mathtools}             
\usepackage{footnote}
\usepackage{tablefootnote}
\usepackage{subfiles}
\usepackage{colortbl}              
\usepackage{wrapfig}               
\usepackage{amssymb}               
\usepackage{comment}               
\usepackage[capitalize]{cleveref}              
\usepackage{pifont}
\usepackage{makecell}
\usepackage[stable,multiple]{footmisc}
\usepackage{bold-extra}
\usepackage{cuted}
\usepackage{pifont}

\usepackage{nicefrac}       
\usepackage{multirow}
\usepackage{cleveref}
\usepackage{dsfont}
\usepackage{makecell}
\usepackage{subcaption}
\usepackage{bbm}
\usepackage{wrapfig}
\usepackage{mathtools,xparse}
\usepackage{array}
\usepackage{arydshln}
\usepackage{scalerel}
\usepackage{tablefootnote}
\usepackage{tocloft}
\usepackage{adjustbox}
\usepackage{setspace}
\usepackage[most,skins,theorems]{tcolorbox}

\usepackage[capitalize]{cleveref}

\definecolor{darkergreen}{RGB}{13, 92, 34}
\definecolor{red2}{RGB}{196, 42, 51}

\newcommand{\yesmark}{\textcolor{darkergreen}{\ding{52}}}
\newcommand{\nomark}{\textcolor{red2}{\ding{56}}}

\newtcolorbox{promptbox}{
  colback=gray!5,
  colframe=gray!70,
  boxrule=0.8pt,
  arc=2mm,
  left=10pt,
  right=10pt,
  top=10pt,
  bottom=10pt,
  width=\linewidth,
  fontupper=\ttfamily\footnotesize,
  enhanced,
}

\newboolean{showsection}
\setboolean{showsection}{true}
\makeatletter
\@namedef{ver@everyshi.sty}{}
\makeatother

\definecolor{custom_green}{rgb}{0.0, 0.5, 0.0}
\definecolor{custom_red}{rgb}{1.0, 0.01, 0.24}
\definecolor{custom_blue}{HTML}{C9DAF7}
\definecolor{custom_purple}{HTML}{D9D1E9}
\definecolor{title_blue}{HTML}{204899}
\definecolor{cite_blue}{HTML}{044dc1}
\definecolor{cite_purple}{HTML}{7406a7}

\hypersetup{
    colorlinks = true,
    citecolor = {cite_blue},
    linkcolor = {cite_purple},
    urlcolor = {cite_purple},
}

\definecolor{blanchedalmond}{rgb}{1.0, 0.92, 0.8}
\definecolor{carmine}{rgb}{0.59, 0.0, 0.09}
\definecolor{lightblue}{rgb}{0.22,0.45,0.70}%

\renewcommand{\mathbf}{\boldsymbol}

\makeatletter
\def\Ddots{\mathinner{\mkern1mu\raise\p@
\vbox{\kern7\p@\hbox{.}}\mkern2mu
\raise4\p@\hbox{.}\mkern2mu\raise7\p@\hbox{.}\mkern1mu}}
\makeatother

\numberwithin{equation}{section}

\definecolor{amaranth}{rgb}{0.9, 0.17, 0.31}
\definecolor{antiquebrass}{rgb}{0.8, 0.58, 0.46}
\definecolor{antiquefuchsia}{rgb}{0.57, 0.36, 0.51}
\definecolor{chromeyellow}{rgb}{0.31, 0.47, 0.26}

\newcommand{\1}{\mathds 1}

\usepackage{amsmath,amsfonts,bm}

\def\eqref#1{equation~\ref{#1}}

\def\1{\bm{1}}

\DeclareMathAlphabet{\mathsfit}{\encodingdefault}{\sfdefault}{m}{sl}
\SetMathAlphabet{\mathsfit}{bold}{\encodingdefault}{\sfdefault}{bx}{n}

\makeatletter
\def\mathcolor#1#{\@mathcolor{#1}}
\def\@mathcolor#1#2#3{%
  \protect\leavevmode
  \begingroup
    \color#1{#2}#3%
  \endgroup
}
\makeatother

\Crefformat{equation}{#2Eq.\;(#1)#3}
\Crefformat{figure}{#2Figure #1#3}
\Crefformat{assumption}{#2Assumption #1#3}
\Crefname{assumption}{Assumption}{Assumptions}

\usepackage{crossreftools}
\makeatletter
\renewcommand\footnoterule{%
  \kern 15\p@
  \hrule \@width 2in \kern 2.6\p@
  \vspace{4pt}
}
\makeatother

\title{Alt-UAV: Altitude-Aware Agent for UAV Navigation}

\reportnumber{}

\author[1,\footnotesize{\textbf{*}}]{Wonjeong Ryu}
\author[1,\footnotesize{\textbf{*}}]{Cholmin Kang}
\author[1]{Junyong Kang}
\author[1,\footnotesize{$\dagger$}]{Hyunjung Shim}

\affil[1]{KAIST AI}

\begin{abstract}
Language-guided UAV target search requires locating potential targets and verifying that they match the instruction.
High-altitude observations provide broad coverage at the expense of object scale, and low-altitude observations offer more detail but are more vulnerable to occlusion.
We propose Alt-UAV, a training-free single-UAV agent that plans observation altitudes and viewpoints for search and verification.
For search, it uses an online 3D global map to select an altitude that meets a same-altitude relative coverage requirement without unnecessary ascent.
When a distant detection is selected as a candidate, Alt-UAV initiates inspection and jointly plans altitude and horizontal position under distance, viewing-angle, and map-based visibility constraints.
The UAV then lands only after a vision-language model confirms the candidate in a close-range image.
On TravelUAV, Alt-UAV outperforms the training-free baseline in success rate across both seen and unseen settings.
Controlled comparisons further show the benefit of search-altitude planning when candidate inspection and close-range confirmation are retained.
\end{abstract}

\begin{document}

{\let\abscontent\relax\maketitle}
\begin{center}
    \centering
    \includegraphics[width=\textwidth]{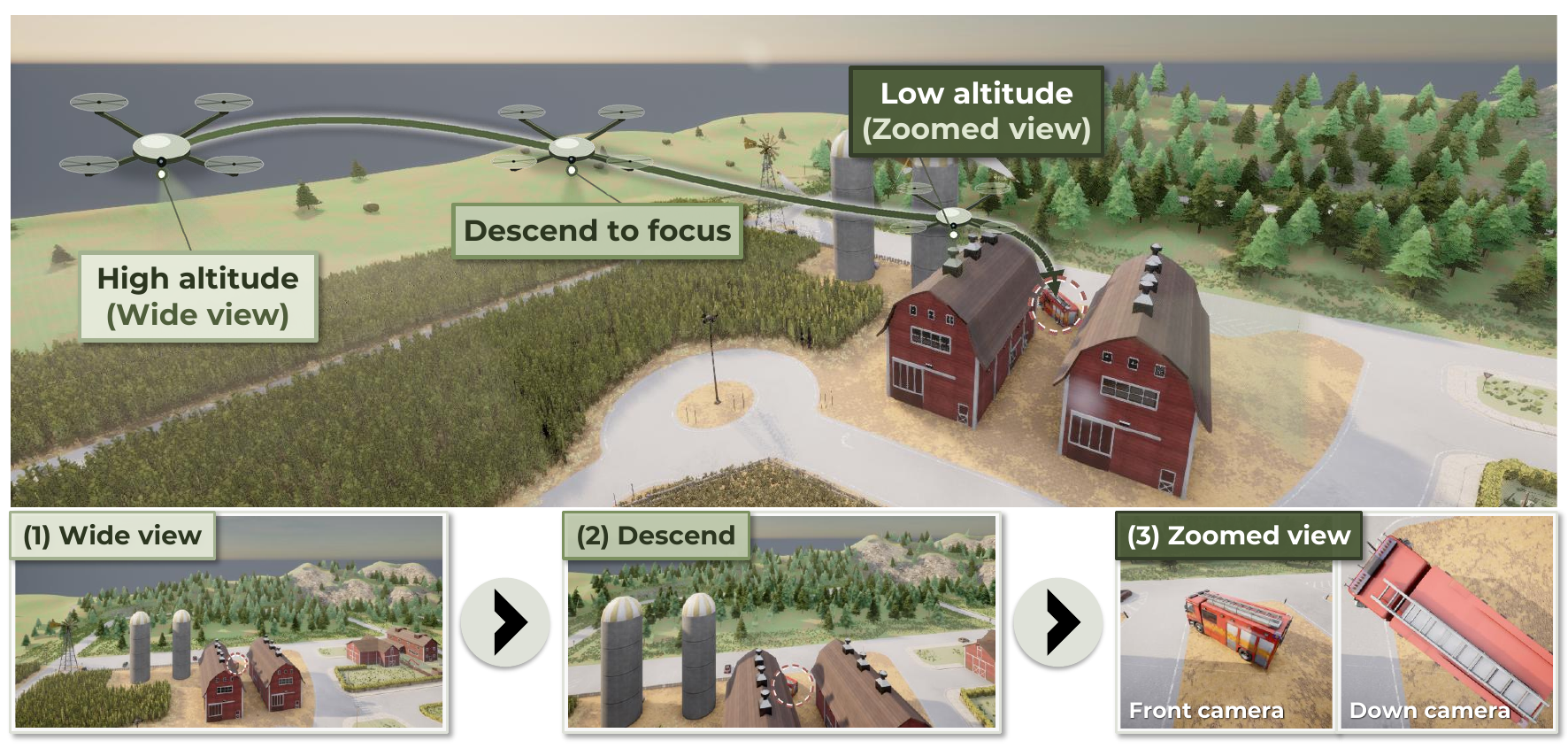}
    \captionsetup{hypcap=false}
    \captionof{figure}{Alt-UAV adjusts its altitude to search above surrounding structures and obtain close-range observations for target verification.}
    \label{fig:teaser}
    \captionsetup{hypcap=true}
\end{center}
\abscontent

\section{Introduction}
\label{sec:intro}
In language-guided target search, a UAV must explore its surroundings and judge whether an object it finds matches the description in the instruction. Unlike recognition on a given image, this task requires the UAV to choose where to observe from, which determines whether the target is visible and recognizable. We call an observation \emph{decisive evidence} when it allows the appearance and relevant spatial relations of a candidate to be checked against the instruction. Since acquiring such evidence depends on visibility, apparent object scale, and viewpoint, the core problem in UAV target search is to choose observations that satisfy these conditions for target verification.

A UAV controls its observations through horizontal motion and altitude. Many UAV search methods select the next region or viewpoint from semantic 3D maps~\citep{Ji2026PRPSearcher,Chen2026AirHunt} and keep altitude fixed or set it only as part of that selection. However, altitude directly affects three requirements of target observation. First, it determines whether the view clears surrounding structures~(visibility). Second, it sets a minimum distance to the ground, which limits how large objects can appear~(scale). Third, at a given horizontal distance, it sets the depression angle, which determines whether a target is seen from the side or from above~(viewpoint).

Choosing an altitude involves a trade-off. Climbing clears structures but reduces apparent object scale, and descending increases scale at the risk of occlusion. Prior work on UAV target search and environmental monitoring addresses this trade-off with precomputed models that relate altitude to recognition accuracy for fixed target classes~\citep{Meera2019OAIPP,stache2023adaptive}. In language-guided search, we do not have such calibrated models for every free-form description, so Alt-UAV plans observation altitudes online from 3D map geometry instead. To avoid choosing a single altitude in this setting, AeroDuo places one UAV at a high altitude and another at a low altitude for detailed search~\citep{Wu2025AeroDuo}. Because using multiple UAVs imposes additional cost, communication, and operational constraints~\citep{FAA107}, a single UAV performing both roles must alternate between climbing to survey and descending to inspect, which requires deciding both when to change altitude and how high to fly.


\begin{figure}[t]
\centering
\captionsetup[sub]{labelsep=space, justification=centering, skip=3pt}
\begin{subfigure}[b]{0.52\linewidth}
    \centering
    \includegraphics[width=\linewidth]{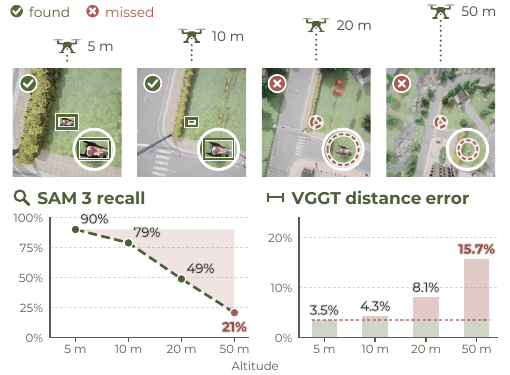}
    \caption{}
    \label{fig:teaser_perc}
\end{subfigure}
\hfill
\begin{subfigure}[b]{0.46\linewidth}
    \centering
    \includegraphics[width=\linewidth]{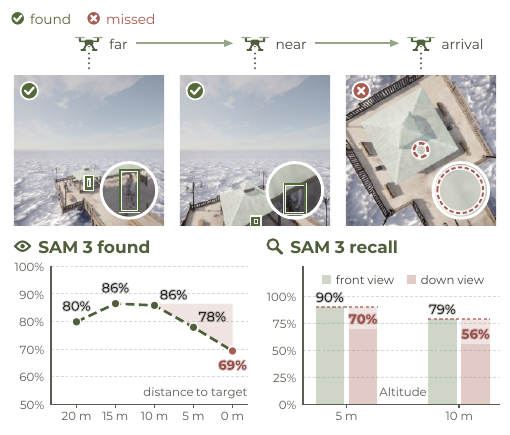}
    \caption{}
    \label{fig:teaser_find}
\end{subfigure}
\caption{(a)~Analysis of recognition performance as a function of altitude. The left graph shows SAM 3~\citep{carion2026sam} performance at different altitudes when prompted with the target. The right graph shows VGGT-based~\citep{wang2025vggt} 3D lifting errors for pixels corresponding to the target. In both graphs, the x-axis is the altitude above the target. And y-axis is the recall in the left graph and the relative distance error in the right graph. This analysis uses the same samples described in \cref{sec:exp_abl}. (b)~Analysis of recognition performance when flying to the target. The left graph shows the fraction of targets that SAM 3~\citep{carion2026sam} detects at each distance to the target, among the targets detected at least once on the way. The right graph shows SAM 3 recall for the front view and the down view taken at the same altitude. The x-axis is the horizontal distance to the target in the left graph and the altitude in the right graph. And y-axis is the detected fraction in the left graph and the recall in the right graph. This analysis uses the same samples described in \cref{sec:exp_abl}.}
\label{fig:teaser_analysis}
\end{figure}

This decision is challenging because each altitude-dependent property causes a distinct perception failure, and changing altitude to avoid one failure leads to the next in a cycle. \textbf{Visibility.} At low altitude, tall structures such as buildings and trees can block the target. \textbf{Scale.} Climbing higher than needed to clear these structures makes objects appear small, which degrades both detection~\citep{carion2026sam} and 3D localization~\citep{wang2025vggt}~(\cref{fig:teaser_perc}). At long range, the agent can miss targets or accept a distant false positive as confirmation and terminate the search at the wrong object. \textbf{Viewpoint.} Approaching a candidate at the search altitude places the UAV directly overhead, where the target leaves the view of the front camera and is often missed in the downward view~(\cref{fig:teaser_find}).  Descending to obtain a side view brings occlusion back. Recent language-guided aerial agents use VLM judgments to adjust altitude~\citep{tong2026visa,Song2026SoraNav} or score exploratory views by coverage gain~\citep{Chen2026AirHunt,Song2026SoraNav}. VLM-based altitude decisions are vulnerable to spatial reasoning errors~\citep{yang2025thinking}, thus an ascent can still leave the scene occluded or shrink objects without a useful gain in coverage. Coverage gain favors newly revealed regions, but a view that exposes more space can still lack the detail and viewing-angle needed to confirm target identity.
This motivates us to compare candidate altitudes and plan inspection poses on a 3D map before moving, and to use the VLM to judge target identity from the acquired images.

We propose a training-free single-UAV agent called Alt-UAV that plans search altitude and candidate inspection on an online 3D map. \textbf{(1)~Search altitude}. Alt-UAV compares predicted surface coverage with a ground-only reference at the same altitude. It selects an altitude that meets this coverage requirement without unnecessary ascent. The coverage requirement guides search based on exposed surfaces on the map. The decline in recognition performance at higher altitudes motivates the limit on ascent~(\cref{fig:teaser_perc}). This geometric policy does not require a target-specific recognition model. \textbf{(2)~Candidate inspection}. Alt-UAV jointly selects altitude and horizontal position to plan an observation pose around each candidate.
It enforces distance and viewing-angle bounds together with a map-based line-of-sight constraint. The agent selects a feasible pose by motion cost and replans as new observations update the map and candidate position. \textbf{(3)~Target confirmation.} Positive or uncertain judgments from a distance initiate inspection rather than landing. The VLM checks whether the candidate matches the instruction in the new close-range image and the UAV lands only after a positive judgment.\\\\
Our contributions are summarized as follows:
\begin{itemize}
    \item We analyze how altitude links visibility, apparent object scale, and viewpoint in language-guided single-UAV target search.
    
    \item We propose a search-altitude planning that couples a same altitude relative coverage requirement with altitude selection to avoid unnecessary ascent.

    \item Our training-free agent plans candidate inspection under distance, viewing-angle, and map-based visibility constraints and requires VLM confirmation from the resulting close-range image before landing.
\end{itemize}

\section{Related Work}
\label{sec:related_work}
\subsection{Language-Guided UAV Navigation}
\label{sec:rel_UAV}
Surveys on UAVs cover control~\citep{mahony2012multirotor}, motion planning~\citep{goerzen2010survey}, computer vision~\citep{kanellakis2017survey}, and the future of small drones~\citep{kumar2012opportunities,floreano2015science}.
Learned control now matches human racing champions~\citep{kaufmann2023champion}, and camera-equipped UAVs have long supported search and rescue~\citep{doherty2007uav,goodrich2008supporting,lin2009uav,waharte2010supporting}.
More recently, natural language has guided UAVs in route following~\citep{Blukis2018Quadcopter,Liu2023AerialVLN}, assisted object search~\citep{Wang2025OpenUAV}, and dialog navigation~\citep{Fan2023AVDN}.
When the goal is an object described by nearby landmarks~\citep{Lee2025CityNav} or by its appearance~\citep{Xiao2025UAVON}, the agent must decide where to explore and whether a candidate is the target.
Learned policies make both decisions from training data.
Such policies generalize across tasks and embodiments~\citep{Zhang2026NavFoM}, handle long flights~\citep{Jiang2025LongFly,Tang2026NeuroKalman}, adapt visual detail by search stage~\citep{Tang2026FreqNav}, and decide when to land~\citep{Xu2026AeroVLA}.
Post-training on failed tasks~\citep{Wang2026RefineFly} or collision-avoidance preferences~\citep{Xu2026AeroDPO} further improves their success rate at the cost of additional training.

Training-free agents avoid this cost by querying pretrained LLMs and VLMs~\citep{achiam2023gpt,team2023gemini,liu2023visual,chen2024internvl,hurst2024gpt,OpenAI2026GPT6Astra,Anthropic2026Opus55} and turning their answers into motion.
For example, an LLM reasons over a top-down map of landmarks~\citep{Gao2025STMR}, or a VLM points to where to fly in the image~\citep{Hu2025SPF,Xu2026Fly0,Ding2026UniLaViRA}.
PRPSearcher and SoraNav also remember explored space to guide the search~\citep{Ji2026PRPSearcher,Song2026SoraNav}.
Recognizing the target, however, depends on the view.
On TravelUAV~\citep{Wang2025OpenUAV} a UAV flying at high altitude can pass over the target without detecting it~\citep{Ding2026UniLaViRA}.
Alt-UAV plans its altitude and viewpoint on a 3D map, and the VLM selects candidates to inspect and judges them from the resulting views.

\subsection{Altitude-Aware Observation Planning}
\label{sec:rel_alt}
A higher altitude provides a wider and less occluded view but makes objects appear smaller.
Path planners handle this trade-off with altitude-dependent models of sensing accuracy~\citep{Meera2019OAIPP,stache2023adaptive,carpin2013variable}, and AeroDuo splits search and close inspection between a high and a low UAV~\citep{Wu2025AeroDuo}.
However, a single UAV must perform both roles, and the higher it searches, the smaller the target appears.
In FlySearch, VLM agents indeed succeed less often when they start higher~\citep{Pardyl2025FlySearch}.
Recent agents therefore adjust altitude to the target size~\citep{Xiao2026AeroBelief} or, like ViSA, in 10m steps~\citep{tong2026visa}.
ViSA notes that altitude changes alone leave facades partly hidden without sideways moves or camera tilt~\citep{tong2026visa}, which motivates planning altitude and viewpoint simultaneously.

Choosing an altitude and viewpoint means deciding where to observe from, which is a core question of active perception~\citep{Bajcsy1988ActivePerception,Bajcsy2018Revisiting}.
Next-best-view planning addresses this question by predicting what each candidate view reveals in a partial scene model~\citep{connolly1985determination,maver1993occlusions,scott2003view,bircher2016receding,strand2025enhancing}.
Object-search agents face the same question in both exploration and verification.
To explore, they rank frontiers~\citep{yamauchi1997frontier} using language-grounded values~\citep{Yokoyama2024VLFM} or Voronoi graphs~\citep{Wu2024VoroNav}, or select aerial viewpoints that surround frontiers and candidates~\citep{Gao2026ImageSpecified} or see promising frontiers~\citep{Chen2026AirHunt}.
To verify candidates, they move closer~\citep{Lei2024IEVE}, combine multiple views~\citep{Yin2024SGNav}, or check the surrounding region against dialog cues~\citep{Qi2026PSCAVDN}.
OA-IPP~\citep{Meera2019OAIPP} combines target-belief maps with an altitude-dependent detector model and AeroBelief~\citep{Xiao2026AeroBelief} adjusts altitude by target size. \citep{Gao2026ImageSpecified} generate viewpoints around frontiers and potential targets and use a learned policy to select action nodes. In contrast, Alt-UAV selects an inspection pose by motion-action cost only after distance, viewing-angle, and map-based line-of-sight constraints are met. \citep{strand2025enhancing} optimize camera pose under visibility and geometry objectives where visibility weights favor less observed surfaces. Alt-UAV uses a ground-only reference at each candidate altitude to define a relative surface coverage requirement for search. It selects an altitude that meets this requirement without unnecessary ascent. Once a candidate is selected, inspection plans a close-range pose under distance, viewing-angle, and map-based visibility constraints, and the VLM checks the resulting image against the instruction before landing.
\begin{figure*}[!t]
\centering
\includegraphics[width=\linewidth]{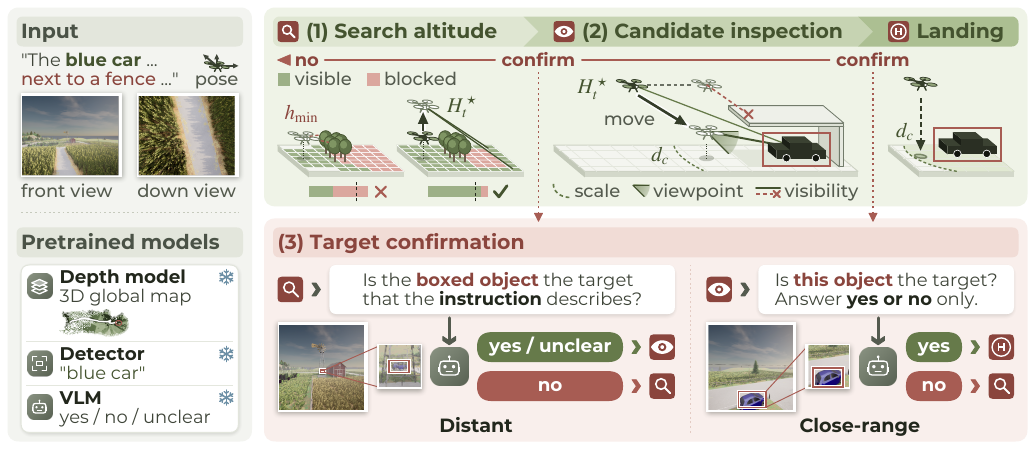}
\caption{Overview of Alt-UAV. Before each move, Alt-UAV casts rays on the online 3D global map to predict the views from candidate poses (hollow UAVs), and then moves from its current pose (solid UAV) toward a pose whose predicted view satisfies the current stage. (1) For the search altitude, the UAV climbs from $h_{\min}$ to $H_t^{\star}$, the lowest altitude at which the predicted coverage relative to a ground-only reference meets the threshold, so it avoids excess ascent and the resulting loss of object scale. (2) For candidate inspection, because an approach at the search altitude would leave the candidate blocked by the roof, the UAV selects a 3D pose from which the candidate lies within $d_c$ (scale), inside the vertical field of view of the front camera (viewpoint), and unoccluded on the map (visibility). (3) For target confirmation, the VLM judges the candidate at each stage transition, where a distant \emph{yes} or \emph{unclear} initiates inspection, a close-range \emph{yes}~(within $d_c$) permits landing, and a close-range no rejects the candidate and returns the UAV to search.}
\label{fig:overview}
\end{figure*}

\section{Method}
\label{sec:meth}

\noindent\textbf{Overview.}
Without task-specific training, Alt-UAV takes instructions, front/down camera views, and pose~$\mathbf{x}_t$ to output motion or landing actions. Instead of learned policies or direct VLM action queries, it casts rays on an online 3D map to evaluate surface coverage during search and line of sight to candidates during inspection, while the VLM judges target identity from the acquired camera images.

\algrenewcommand{\algorithmiccomment}[1]{\hfill(#1)}

\begin{wrapfigure}{R}{0.5\textwidth}
\vspace{-\intextsep}
\begin{minipage}{\linewidth}
\begin{algorithm}[H]
\caption{Alt-UAV}
\label{alg:altuav}
\small
\begin{algorithmic}[1]
\State \textbf{Inputs:} instruction~$\mathcal{I}$, depth model~\citep{wang2025vggt}, detector~\citep{carion2026sam}, VLM~\citep{bai2025qwen3}
\For{each step~$t$}
  \State Update the 3D global map~$\mathcal{M}_t$ \Comment{depth model}
  \State Detect objects that match the target \Comment{detector}
  \If{target confirmed at close range} \Comment{\cref{sec:meth_evid}}
    \State \Return $\mathrm{LAND}$
  \ElsIf{a candidate is selected} \Comment{\cref{sec:meth_evid}}
    \State Plan a move to an inspection pose \Comment{\cref{sec:meth_insp}}
  \Else
    \State $H^\star_t \gets$ search altitude on $\mathcal{M}_t$ \Comment{\cref{eq:hstar}}
    \State Plan a move at altitude $H^\star_t$ \Comment{\cref{sec:meth_search}}
  \EndIf
  \State Execute the move as a motion action
\EndFor
\end{algorithmic}
\end{algorithm}
\end{minipage}
\end{wrapfigure}
Alt-UAV operates in three stages as described in \cref{fig:overview} and \cref{alg:altuav}. First, it selects a search altitude that meets a relative coverage requirement without unnecessary ascent~(\cref{sec:meth_search}). Second, it inspects candidates under distance, angle, and visibility constraints~(\cref{sec:meth_insp}). Third, it verifies whether the candidate matches the instruction in a close-range image before landing~(\cref{sec:meth_evid}).
All three stages share three pretrained modules operating solely on the two camera views and pose $\mathbf{x}_t$. \textbf{Map module:} maintains a metric voxel occupancy map~$\mathcal{M}_t$~\citep{tian2023occ3d} for planning by accumulating predicted depths~\citep{wang2025vggt}  transformed via $\mathbf{x}_t$. \textbf{Detection module:} uses SAM~3~\citep{carion2026sam} to segment target-named objects, projecting each object~$i$'s mask center along its ray to the first occupied voxel to obtain 3D position~$P_i$ and UAV distance~$d_i$.
Detections are classified as close-range if  $d_i\le d_c$  and distant otherwise, with the confirmation threshold~$d_c$ set below the task success radius~$r_{\mathrm{succ}}$. \textbf{Judgment module:} queries the VLM~(Qwen3-VL-Instruct~\citep{bai2025qwen3})  to extract the target name for detection, judge whether boxed object~$i$ matches the instruction ($j_i \in \{\text{\textit{yes}}, \text{\textit{no}}, \text{\textit{unclear}}\}$), or point to the target when detection fails. Given the VLM's limited spatial reasoning~\citep{yang2025thinking}, it outputs neither altitudes nor actions.
The action space comprises motion actions~$(d^{\mathrm{fwd}}_t,d^{\mathrm{dn}}_t,\Delta\psi_t)$ and $\mathrm{LAND}$. The UAV executes $\mathrm{LAND}$ only upon target confirmation~(\cref{sec:meth_evid}). Otherwise, the UAV moves toward planned inspection poses or search altitudes, exploring horizontally via frontier-based exploration~\citep{yamauchi1997frontier} along the instructed direction during the search.

\subsection{Search Altitude}
\label{sec:meth_search}
We compare the horizontal extent of visible mapped surfaces against a ground-only reference at the same altitude, and select an altitude that meets the coverage requirement without unnecessary ascent.

Let $\mathcal{H}$ be a set of altitude levels between $h_{\min}$ and $h_{\max}$, measured relative to the surface beneath the UAV by triangulation from the down images.
For each level $h\in\mathcal{H}$ and front or down view $v\in\{\mathrm{F},\mathrm{D}\}$, we cast rays from a camera at altitude~$h$ onto a map~$\mathcal{M}$.
The resulting coverage~$C_v(h;\mathcal{M})$ counts the horizontal grid cells hit within range~$r_{\max}$ and measures the extent of the visible surfaces such as the ground and the tops of structures.
Since the camera footprint expands with altitude even on flat ground, we compare the coverage predicted on the current map~$\mathcal{M}_t$ with the coverage on a ground-only reference map~$\mathcal{M}_{\varnothing}$ at the same altitude:
\begin{equation}
\label{eq:rho}
\rho_v(h)=
\frac{C_v(h;\mathcal{M}_t)}{C_v(h;\mathcal{M}_{\varnothing})},
\end{equation}
where the reference coverage is assumed positive. Because the two maps have different visible cells, this ratio compares coverage extents rather than cell-wise overlap. In the current map $\mathcal{M}_t$, the numerator~$C_v(h;\mathcal{M}_t)$ measures the horizontal spread of all visible surfaces, such as the ground and rooftops, whereas the denominator~$C_v(h;\mathcal{M}_\varnothing)$ gives the reference footprint for the same camera view at altitude $h$. This ground-only reference accounts for how the camera footprint changes with altitude rather than assuming that targets lie on the ground. The numerator counts a rooftop hit as coverage even though the ground beneath that roof remains hidden. Before a candidate is found, this relative surface coverage guides the search altitude. Because this rule relies solely on mapped 3D geometry rather than predicted detection accuracy, it requires no target-specific recognition profile. Once a candidate is selected, the UAV shifts from this scene-level search criterion to close-range viewpoint planning under distance, viewing-angle, and map-based line-of-sight constraints~(\cref{sec:meth_insp}). We select the search altitude at which the coverage ratio in both views meets the threshold $1-\epsilon$:
\begin{equation}
\label{eq:hstar}
H^\star_t=
\min\left\{
h\in\mathcal{H}\ \middle|\
\begin{aligned}
&\rho_v(h)\ge 1-\epsilon,\\
&\forall v\in\{\mathrm F,\mathrm D\}
\end{aligned}
\right\}.
\end{equation}
We apply this threshold to both views because the detector runs on both images. A smaller $\epsilon$ demands greater surface coverage relative to the flat-ground reference, and taking the minimum altitude in \cref{eq:hstar} satisfies this requirement without climbing higher than needed and shrinking object scale.

Recomputing~$H^\star_t$ on the updated map at every step, the UAV changes its altitude one level at a time, climbing whenever below~$H^\star_t$ but descending only when more than one level above it, to avoid oscillation.
If no level in $\mathcal{H}$ satisfies \cref{eq:hstar}, the UAV relocates to a position from which it can safely climb when the space ahead is blocked. Otherwise, it descends one level if the lower path is clear, or holds altitude.

\subsection{Candidate Inspection}
\label{sec:meth_insp}

Inspection begins upon selecting a candidate, a distant detection judged as either \textit{yes} or \textit{unclear}, or a VLM-pointed pixel mapped like a mask center when detection fails. This inclusive selection accommodates small distant objects and missed detections. Since candidates are unconfirmed, the UAV inspects them one at a time at position~$P$ to obtain decisive evidence before landing.

A direct approach to $P$ does not guarantee a suitable view for verification. Overhead views may lose or miss $P$ across front or down cameras~(\cref{fig:teaser_find}), and nearby structures cause occlusion. Once inspection begins, the UAV jointly selects altitude and horizontal position around $P$ under distance, viewing-angle, and map-based visibility constraints.

For a pose~$\mathbf{x}$ (position and heading), let $d(\mathbf{x})$ and $\theta(\mathbf{x})$ be the length and downward angle of the line of sight from the front camera at $\mathbf{x}$ to $P$.
We cast this line of sight onto $\mathcal{M}_t$ as in search and set $V_t(\mathbf{x})=1$ if it passes through no intervening occupied voxel before reaching the candidate voxel, and $V_t(\mathbf{x})=0$ otherwise.
The inspection poses are those that satisfy the three geometric constraints:
\begin{equation}
\label{eq:inspection_set}
\mathcal{Q}_t(P)=
\left\{\mathbf{x}\ \middle|\
\begin{array}{@{}l@{\quad}l@{}}
d(\mathbf{x})\le d_c & \text{(scale)}\\
\theta(\mathbf{x})<\varphi/2 & \text{(viewpoint)}\\
V_t(\mathbf{x})=1 & \text{(visibility)}
\end{array}
\right\},
\end{equation}
where $\varphi$ is the vertical field of view of the front camera.
The distance bound brings the UAV within a fixed range~$d_c$ across all targets so that the VLM judges the candidate from a close-range image.
The angle bound keeps the candidate within this field of view, while the visibility test excludes map-predicted occlusions.

Together, the distance and angle bounds keep the pose within $d_c\sin(\varphi/2)$ above $P$ and push higher poses farther back, setting altitude relative to the candidate rather than surrounding structures. From $\mathcal{Q}_t(P)$, the UAV selects the goal pose requiring the fewest motion actions:
\begin{equation}
\label{eq:inspection_plan}
\mathbf{x}^\star_t=\operatorname*{arg\,min}_{\mathbf{x}\in\mathcal{Q}_t(P)} c_t(\mathbf{x}),
\end{equation}
where $c_t(\mathbf{x})$ counts the motion actions needed to reach $\mathbf{x}$ without collision on $\mathcal{M}_t$.
While the constraints in $\mathcal{Q}_t(P)$ define feasible close-range poses on the current map, the UAV selects the one with the smallest $c_t(\mathbf{x})$, and the VLM then checks the resulting close-range image against the instruction~(\cref{sec:meth_evid}).

As the UAV approaches the candidate, new observations update both the estimated position~$P$ and the map $\mathcal{M}_t$. After every motion action, we replan the inspection pose on the updated map~\citep{bircher2016receding} and choose actions that keep the candidate in the front camera view.

\subsection{Target Confirmation}
\label{sec:meth_evid}
The last stage decides whether to land. Landing requires stronger evidence than starting an inspection because it ends the episode and cannot be undone, and judging a target from a distance is prone to error~\citep{Pardyl2025FlySearch}.
A distant positive or uncertain judgment initiates inspection rather than landing. As the UAV moves toward the planned close-range pose~(\cref{sec:meth_insp}), it evaluates the candidate at every step and defers the final decision until a close-range view is acquired.

At every step, a new detection~$i$ placed near~$P$ on the map is taken as the candidate and updates~$P$.
Once the candidate is at close range, the VLM judges it again with only \textit{yes} or \textit{no}, and the decision on this detection~$i$ at each step is
\begin{equation}
\label{eq:confirm}
\begin{cases}
\mathrm{LAND} & \text{if } d_i\le d_c \text{ and } j_i=\textit{yes},\\
\text{reject} & \text{if } d_i\le d_c \text{ and } j_i=\textit{no},\\
\text{defer} & \text{otherwise}.
\end{cases}
\end{equation}
Only a close-range observation thus ends the inspection with a decision, and the same \textit{yes} that merely starts an inspection from afar permits a landing at close range.
In this way, Alt-UAV uses uncertain distant cues to guide inspection and requires a positive close-range judgment before landing.
To avoid repeated inspections within the action budget, the UAV ignores any candidate rejected at close range for the rest of the episode and returns to the search altitude~(\cref{sec:meth_search}) to resume searching. If a detected candidate disappears for several consecutive steps, the UAV resumes the search without a decision. A candidate location pointed to by the VLM is approached like any other candidate. It becomes a detected candidate once an object is detected there and is rejected if none appears even at close range. A detection that is already within~$d_c$ when first seen in the search stage needs no approach and is decided by \cref{eq:confirm} at once.

\section{Experiments}
\label{sec:exp}

\subsection{Experimental Setup}
\label{sec:exp_set}

\begin{table*}[!t]
\caption{Quantitative comparison on the Test Seen split across Full, Easy, and Hard evaluation sets. $^\dagger$ indicates methods based on AeroVLA, which introduces fuzzy direction guidance to provide the relative direction to the target at each navigation step. We evaluate PRPSearcher with collisions ignored, following its released implementation~\citep{Ji2026PRPSearcher}. The best results are in \textbf{bold}, and the second-best results are \underline{underlined}.}
\label{tab:uav_nav}
\centering
\footnotesize
\setlength{\tabcolsep}{2pt}
\renewcommand{\arraystretch}{0.97}
\setlength{\aboverulesep}{1pt}
\setlength{\belowrulesep}{1pt}
\begin{tabularx}{\textwidth}{@{}l*{4}{>{\centering\arraybackslash}X}|*{4}{>{\centering\arraybackslash}X}|*{4}{>{\centering\arraybackslash}X}@{}}
\toprule
\multirow{2}{*}[-0.5ex]{Method}
& \multicolumn{4}{c|}{Full}
& \multicolumn{4}{c|}{Easy}
& \multicolumn{4}{c}{Hard} \\
\cmidrule(lr){2-5}\cmidrule(lr){6-9}\cmidrule(lr){10-13}
& NE$\downarrow$ & SR$\uparrow$ & OSR$\uparrow$ & SPL$\uparrow$
& NE$\downarrow$ & SR$\uparrow$ & OSR$\uparrow$ & SPL$\uparrow$
& NE$\downarrow$ & SR$\uparrow$ & OSR$\uparrow$ & SPL$\uparrow$ \\
\specialrule{0.4pt}{1pt}{0pt}
NavFoM~\citep{Zhang2026NavFoM}
& 93.05 & 29.17 & 49.24 & 25.03
& 58.98 & 32.91 & 53.16 & 27.87
& 143.83 & 23.58 & 43.40 & 20.80 \\
NeuroKalman~\citep{Tang2026NeuroKalman}
& 71.56 & 25.86 & 58.73 & 22.43
& 42.70 & 30.52 & 62.70 & 25.86
& 105.07 & 20.11 & 53.90 & 18.21 \\
AeroVLA$^\dagger$~\citep{Xu2026AeroVLA}
& 65.88 & 47.96 & 57.69 & 38.54
& 43.76 & 49.30 & 61.30 & 37.14
& 93.16 & 46.30 & 53.23 & 40.26 \\
FreqNav$^\dagger$~\citep{Tang2026FreqNav}
& 64.16 & 51.55 & 60.79 & 41.67
& 45.64 & 49.81 & 60.54 & 38.04
& 86.99 & 53.70 & 61.10 & 46.14 \\
RefineFly$^\dagger$~\citep{Wang2026RefineFly}
& 54.96 & 56.33 & 65.40 & 45.98
& 37.60 & 56.28 & 67.39 & 42.91
& 76.36 & 56.38 & 62.94 & 49.75 \\
AeroDPO$^\dagger$~\citep{Xu2026AeroDPO}
& 53.30 & \underline{60.93} & 67.49 & \underline{48.47}
& 36.03 & \underline{63.73} & 70.75 & \underline{47.13}
& 74.60 & \underline{57.48} & 63.47 & \underline{50.11} \\
PRPSearcher~\citep{Ji2026PRPSearcher}
& 62.21 & 36.67 & 63.19 & 29.20
& 54.85 & 41.12 & \underline{71.65} & 31.27
& 71.13 & 31.18 & 52.76 & 26.65 \\
\rowcolor{darkergreen!10}
\textbf{Alt-UAV~(Ours)}
& \underline{44.15} & 55.29 & \underline{69.68} & 44.75
& \underline{35.95} & 55.30 & 71.52 & 42.75
& \underline{54.25} & 55.28 & \underline{67.40} & 47.21 \\
\rowcolor{darkergreen!10}
\textbf{Alt-UAV$^\dagger$~(Ours)}
& \textbf{34.92} & \textbf{65.23} & \textbf{76.09} & \textbf{51.86}
& \textbf{23.94} & \textbf{67.94} & \textbf{80.08} & \textbf{51.40}
& \textbf{48.45} & \textbf{61.89} & \textbf{71.18} & \textbf{52.43} \\
\specialrule{0.8pt}{0pt}{0pt}
\end{tabularx}
\end{table*}
\noindent\textbf{Dataset.}
We evaluate Alt-UAV on TravelUAV~\cite{Wang2025OpenUAV}, a target-oriented VLN benchmark in AirSim with 12k instruction-trajectory pairs from 22 outdoor urban, rural, and natural environments. Each instruction describes the target, its relative direction, and its surroundings. The agent must visually search for the target and stop on its own within 200 actions. Whereas AerialVLN~\cite{Liu2023AerialVLN} uses predefined motion primitives and CityNav~\cite{Lee2025CityNav} focuses on geographic grounding, TravelUAV simulates physical 3D flight for visual target search, which allows us to evaluate both altitude changes and stopping at the target. We report results on the Seen, Unseen Object, and Unseen Map splits.

\noindent\textbf{Evaluation Metrics.}
Following the benchmark protocol, we adopt Success Rate~(SR), Oracle Success Rate~(OSR), Navigation Error~(NE), and Success weighted by Path Length~(SPL)~\cite{Anderson2018Evaluation}. All distances are computed in 3D space to account for both horizontal displacement and altitude, and the success threshold is set to 20m~\citep{Wang2025OpenUAV}.

\subsection{Main Results}
\cref{tab:uav_nav,tab:uav_unseen,tab:uav_map} compare Alt-UAV with learning-based navigation policies and the training-free agent PRPSearcher across the Seen, Unseen Object, and Unseen Map splits. To control for pretrained model capacity, we evaluate PRPSearcher~\citep{Ji2026PRPSearcher} with the same pretrained models as Alt-UAV. We also report comparisons against SoraNav~\citep{Song2026SoraNav}, adapted to TravelUAV within our pretrained models, in the supplementary material.

Without fuzzy direction guidance, Alt-UAV achieves 55.29\%, 67.57\%, and 45.09\% SR on the Full sets of Seen, Unseen Object, and Unseen Map, respectively, which surpasses PRPSearcher by 18.62, 33.07, and 8.35 percentage points. It also reaches 44.75\%, 58.46\%, and 34.00\% SPL on the three splits, for gains of 15.55, 29.07, and 4.26 percentage points over PRPSearcher. Because both agents share the exact same pretrained modules and instruction settings, these margins demonstrate that explicit search-altitude and viewpoint planning substantially improves both target discovery and path efficiency across seen and unseen settings.

With fuzzy direction guidance, Alt-UAV achieves the highest SR and SPL among all compared methods across all three Full sets. It reaches 65.23\%, 70.75\%, and 57.83\% SR on Seen, Unseen Object, and Unseen Map, respectively, outperforming the strongest learned policy, AeroDPO, by 4.30, 3.34, and 8.67 percentage points, alongside SPL gains of 3.39, 2.61, and 3.27 percentage points. Without any task-specific training, Alt-UAV surpasses learned navigation policies under the same guidance and achieves its largest lead on Unseen Map, where trained policies degrade in unfamiliar layouts.

\begin{figure*}[!t]
\centering
\includegraphics[width=\linewidth]{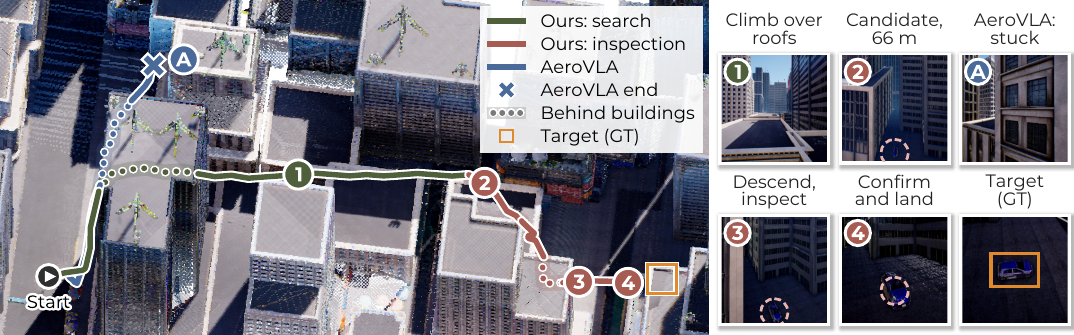}
\vspace{-1.5em}
\caption{
Trajectories and observations of Alt-UAV and AeroVLA. Alt-UAV searches above rooftops and descends for verification, while AeroVLA becomes stuck near a building.
}
\label{fig:main_qual}
\end{figure*}

\begin{table*}[!t]
\caption{Quantitative comparison on the Test Unseen Object split across Full, Easy, and Hard evaluation sets. The experimental setup and notations follow \cref{tab:uav_nav}.}
\vspace{-0.5em}
\label{tab:uav_unseen}
\centering
\footnotesize
\setlength{\tabcolsep}{2pt}
\renewcommand{\arraystretch}{0.97}
\setlength{\aboverulesep}{1pt}
\setlength{\belowrulesep}{1pt}
\begin{tabularx}{\textwidth}{@{}l*{4}{>{\centering\arraybackslash}X}|*{4}{>{\centering\arraybackslash}X}|*{4}{>{\centering\arraybackslash}X}@{}}
\toprule
\multirow{2}{*}[-0.5ex]{Method}
& \multicolumn{4}{c|}{Full}
& \multicolumn{4}{c|}{Easy}
& \multicolumn{4}{c}{Hard} \\
\cmidrule(lr){2-5}\cmidrule(lr){6-9}\cmidrule(lr){10-13}
& NE$\downarrow$ & SR$\uparrow$ & OSR$\uparrow$ & SPL$\uparrow$
& NE$\downarrow$ & SR$\uparrow$ & OSR$\uparrow$ & SPL$\uparrow$
& NE$\downarrow$ & SR$\uparrow$ & OSR$\uparrow$ & SPL$\uparrow$ \\
\specialrule{0.4pt}{1pt}{0pt}

NavFoM~\citep{Zhang2026NavFoM}
& 108.04 & 29.83 & 47.99 & 27.20
& 70.51 & 32.54 & 50.72 & 29.54
& 133.01 & 28.03 & 46.18 & 25.64 \\
NeuroKalman~\citep{Tang2026NeuroKalman}
& 71.01 & 32.48 & 60.82 & 28.50
& 44.50 & 42.50 & 66.50 & 37.37
& 84.50 & 27.50 & 58.00 & 24.50 \\
AeroVLA$^\dagger$~\citep{Xu2026AeroVLA}
& 61.45 & 56.60 & 64.86 & 46.61
& 45.72 & 56.94 & 64.11 & 43.76
& 69.27 & 56.43 & 65.24 & 48.03 \\
FreqNav$^\dagger$~\citep{Tang2026FreqNav}
& 58.39 & 57.23 & 70.75 & 47.00
& 36.96 & 56.94 & 72.73 & 45.02
& 69.06 & 57.38 & 69.76 & 47.99 \\
RefineFly$^\dagger$~\citep{Wang2026RefineFly}
& 53.42 & 59.72 & 68.31 & 51.34
& 35.49 & 63.80 & 72.09 & 50.97
& 62.34 & 57.70 & 66.43 & 51.52 \\
AeroDPO$^\dagger$~\citep{Xu2026AeroDPO}
& 47.31 & 67.41 & 71.55 & 57.54
& 33.31 & 67.47 & 72.73 & 54.27
& 54.27 & \underline{67.38} & 70.95 & \underline{59.16} \\
PRPSearcher~\citep{Ji2026PRPSearcher}
& 59.43 & 34.50 & 56.12 & 29.39 
& 53.48 & 44.50 & 73.21 & 35.37 
& 62.39 & 29.52 & 47.62 & 26.42 \\
\rowcolor{darkergreen!10}
\textbf{Alt-UAV~(Ours)}
& \underline{37.36} & \underline{67.57} & \textbf{79.17} & \underline{58.46}
& \underline{29.60} & \underline{70.33} & \underline{82.78} & \underline{57.40}
& \textbf{41.22} & 66.19 & \textbf{77.38} & 58.99 \\
\rowcolor{darkergreen!10}
\textbf{Alt-UAV$^\dagger$~(Ours)}
& \textbf{35.60} & \textbf{70.75} & \underline{78.86} & \textbf{60.15}
& \textbf{21.41} & \textbf{75.60} & \textbf{83.73} & \textbf{60.86}
& \underline{42.65} & \textbf{68.33} & \underline{76.43} & \textbf{59.80} \\
\specialrule{0.8pt}{0pt}{0pt}
\end{tabularx}
\end{table*}
\begin{table*}[!t]
\vspace{-0.5em}
\caption{Quantitative comparison on the Test Unseen Map split across Full, Easy, and Hard evaluation sets.  The experimental setup and notations follow \cref{tab:uav_nav}.}
\vspace{-0.5em}
\label{tab:uav_map}
\centering
\footnotesize
\setlength{\tabcolsep}{2pt}
\renewcommand{\arraystretch}{0.97}
\setlength{\aboverulesep}{1pt}
\setlength{\belowrulesep}{1pt}
\begin{tabularx}{\textwidth}{@{}l*{4}{>{\centering\arraybackslash}X}|*{4}{>{\centering\arraybackslash}X}|*{4}{>{\centering\arraybackslash}X}@{}}
\toprule
\multirow{2}{*}[-0.5ex]{Method}
& \multicolumn{4}{c|}{Full}
& \multicolumn{4}{c|}{Easy}
& \multicolumn{4}{c}{Hard} \\
\cmidrule(lr){2-5}\cmidrule(lr){6-9}\cmidrule(lr){10-13}
& NE$\downarrow$ & SR$\uparrow$ & OSR$\uparrow$ & SPL$\uparrow$
& NE$\downarrow$ & SR$\uparrow$ & OSR$\uparrow$ & SPL$\uparrow$
& NE$\downarrow$ & SR$\uparrow$ & OSR$\uparrow$ & SPL$\uparrow$ \\
\specialrule{0.4pt}{1pt}{0pt}

NavFoM~\citep{Zhang2026NavFoM}
& 125.10 & 6.30 & 18.95 & 5.68
& 102.41 & 6.77 & 20.07 & 6.04
& 170.58 & 5.36 & 15.71 & 4.97 \\
NeuroKalman~\citep{Tang2026NeuroKalman}
& 100.32 & 8.34 & 34.15 & 7.12
& 69.50 & 9.15 & 38.50 & 7.50
& 140.00 & 7.20 & 28.00 & 6.50 \\
AeroVLA$^\dagger$~\citep{Xu2026AeroVLA}
& 67.42 & 37.58 & 52.92 & 28.22
& 44.99 & 41.89 & 58.47 & 29.72
& 99.11 & 31.49 & 45.09 & 26.11 \\
FreqNav$^\dagger$~\citep{Tang2026FreqNav}
& 72.91 & 39.04 & 55.11 & 30.50
& 53.90 & 40.29 & 58.82 & 29.88
& 99.77 & 37.28 & 49.87 & 31.38 \\
RefineFly$^\dagger$~\citep{Wang2026RefineFly}
& 58.88 & 42.97 & 60.09 & 31.83
& 44.02 & 46.88 & 63.04 & 32.44
& 79.88 & 37.45 & 55.92 & 30.98 \\
AeroDPO$^\dagger$~\citep{Xu2026AeroDPO}
& 66.96 & \underline{49.16} & 57.72 & \underline{39.06}
& 48.70 & \underline{52.23} & 61.32 & \underline{39.48}
& 92.76 & \underline{44.84} & 52.64 & \underline{38.46} \\
PRPSearcher~\citep{Ji2026PRPSearcher}
& 67.54 & 36.74 & \textbf{71.09} & 29.74 
& 57.82 & 39.57 & \underline{75.76} & 30.77
& 81.11 & 32.75 & \textbf{64.48} & 28.28 \\
\rowcolor{darkergreen!10}
\textbf{Alt-UAV~(Ours)}
& \underline{52.31} & 45.09 & 63.78 & 34.00
& \underline{42.86} & 47.95 & 67.20 & 35.35
& \textbf{65.67} & 41.06 & 58.94 & 32.08 \\
\rowcolor{darkergreen!10}
\textbf{Alt-UAV$^\dagger$~(Ours)}
& \textbf{45.45} & \textbf{57.83} & \underline{70.67} & \textbf{42.33}
& \textbf{30.53} & \textbf{62.75} & \textbf{76.65} & \textbf{44.64}
& \underline{66.53} & \textbf{50.88} & \underline{62.22} & \textbf{39.06} \\
\specialrule{0.8pt}{0pt}{0pt}
\end{tabularx}
\end{table*}
As seen in \cref{fig:main_qual}, Alt-UAV climbs above rooftops to search~(1) and detects a candidate 66m away~(2). It then descends to a planned inspection viewpoint~(3) and confirms the target before landing~(4), while AeroVLA stuck near a building~(A). The distant detection guides inspection, while newly acquired close-range evidence determines landing.

\subsection{Ablation Study}
\label{sec:exp_abl}
All ablations were run on 476~episodes sampled from the TravelUAV training split. We grouped the episodes by environment, target size, start-to-target distance, and difficulty. Also, we sampled each group in proportion to its size, so that the episodes include every combination of these attributes that occurs in the training split and follow its distribution. None of the pretrained models were trained or fine-tuned on TravelUAV, and test splits were reserved for main comparisons.

\begin{table}[t]
\begin{minipage}[b]{0.54\textwidth}
\caption{Ablation of search-altitude planning. We compare search-altitude strategies while retaining candidate inspection and close-range confirmation in all variants, without fuzzy direction.}
\label{tab:abl_alt}
\centering
\renewcommand{\arraystretch}{1.0}
\footnotesize\setlength{\tabcolsep}{2pt}
\begin{adjustbox}{max width=\linewidth}
\begin{tabular}{l|cccc}
\toprule
Method & NE$\downarrow$ & SR$\uparrow$ & OSR$\uparrow$ & SPL$\uparrow$\\

\specialrule{0.6pt}{1pt}{1pt}

w/o Search-altitude Planning 
& 102.49 & 21.64 & 33.40 & 17.72 \\
ViSA-inspired~(VLM, $\pm$5m)~\citep{tong2026visa}
& 67.54 & 37.82 & 48.53 & 22.23 \\
Fixed clearance~(5m)
& \underline{57.58} & \underline{41.39} & \underline{61.13} & \underline{30.94} \\

\specialrule{0.6pt}{1pt}{0pt}
\rowcolor{darkergreen!10}
\rule[-4pt]{0pt}{14pt}\textbf{Alt-UAV~(Ours)}
& \textbf{47.98} & \textbf{55.04} & \textbf{69.33} & \textbf{45.71} \\
\specialrule{0.8pt}{0pt}{0pt}
\end{tabular}
\end{adjustbox}
\end{minipage}\hfill
\begin{minipage}[b]{0.42\textwidth}
\caption{Ablation of inspection and candidate selection. \textit{Inspection} plans the altitude and position for viewing a candidate~(\cref{sec:meth_insp}), and \textit{Selection} decides which detections are inspected~(\cref{sec:meth_evid}). The rest of Alt-UAV is the same in every row.}
\label{tab:abl_remain}
\centering
\renewcommand{\arraystretch}{0.92}
\footnotesize\setlength{\tabcolsep}{2pt}
\begin{adjustbox}{max width=\linewidth}
\begin{tabular}{cc|cccc}
\toprule
Inspection & Selection & NE$\downarrow$ & SR$\uparrow$ & OSR$\uparrow$ & SPL$\uparrow$ \\

\specialrule{0.6pt}{1pt}{1pt}

\nomark & \nomark
& \underline{47.68} & 47.90 & 68.70 & 39.67 \\
\nomark & \yesmark
& 50.81 & \underline{51.47} & 67.86 & \underline{42.75} \\
\yesmark & \nomark
& \textbf{42.22} & \underline{51.47} & \textbf{69.33} & 42.55 \\

\specialrule{0.6pt}{1pt}{0pt}
\rowcolor{darkergreen!10}
\rule[-4pt]{0pt}{14pt}\yesmark & \yesmark
& 47.98 & \textbf{55.04} & \textbf{69.33} & \textbf{45.71} \\
\specialrule{0.8pt}{0pt}{0pt}
\end{tabular}
\end{adjustbox}
\end{minipage}

\end{table}
\noindent\textbf{Search-altitude Planning.}
\cref{tab:abl_alt} compares search-altitude strategies while perception, horizontal exploration, candidate inspection, and close-range confirmation remain fixed, without fuzzy direction guidance. Without search-altitude planning, the UAV follows minor terrain changes at its initial ground height and detours around buildings and trees, achieving 21.64\% SR and 33.40\% OSR. The ViSA-inspired baseline~\citep{tong2026visa} selects $+$5m, 0m, or $-$5m from the current views and altitude history at each search step and achieves 37.82\% SR and 22.23 SPL. Fixed clearance uses the 3D map to climb 5m above obstacles on the forward path and returns to 5m above the ground afterward, reaching 41.39\% SR. This rule sets altitude from path obstacles rather than predicted scene coverage. Alt-UAV instead evaluates relative surface coverage in both views and limits further ascent once the requirement is met~(\cref{eq:hstar}). The recognition and localization trends in \cref{fig:teaser_perc} motivate this ascent limit. Alt-UAV achieves 55.04\% SR, 69.33\% OSR, and 45.71 SPL, exceeding fixed clearance by 13.65 percentage points in SR, 8.20 points in OSR, and 14.77 points in SPL. These gains support selecting search altitude for observation rather than path clearance alone.

\noindent\textbf{Inspection \& Selection.}
\cref{tab:abl_remain} evaluates candidate inspection and candidate selection while search-altitude planning and close-range confirmation~(\cref{eq:confirm}) remain fixed. Selection queries the VLM before approach to filter out distant detections that do not match the instruction. Without inspection, however, the UAV flies directly toward the selected candidate at the search altitude rather than a planned close-range pose, so even a true target can become occluded or leave the front view as the UAV nears it~(\cref{fig:teaser_find}). This geometric limitation is consistent with the higher NE~(50.81 vs. 42.22) and lower OSR~(67.86\% vs. 69.33\%) of selection alone relative to inspection alone.

Inspection plans close-range viewpoints under distance, viewing-angle, and map-based visibility constraints in \cref{eq:inspection_set} to keep the candidate observable during approach, which achieves the lowest NE~(42.22) and highest OSR~ (69.33\%). Even with unobstructed views, however, SR plateaus at 51.47\% because a detector built for short noun phrases only matches object categories and cannot verify full descriptions or relations across the scene. Before approach, the agent still requires the VLM to evaluate the candidate within the broader scene context so that it does not pursue every category match as a valid target.

While each component alone stops at 51.47\% SR, their combination reaches 55.04\% SR and 45.71 SPL. Scene-level VLM selection is effective only when the subsequent approach preserves a clear view of the chosen candidate, and geometric viewpoint planning translates into successful confirmation only when directed at candidates that satisfy the instruction relations across the scene.
\section{Conclusion}
\label{sec:con}
We presented a training-free single-UAV agent called Alt-UAV for language-guided target search. Its search-altitude policy couples a same-altitude relative coverage requirement with an altitude selection rule that avoids unnecessary ascent. The agent plans candidate inspection poses under distance and viewing-angle bounds with a map-based visibility constraint and lands only after a positive VLM judgment on a new close-range image. Experiments on TravelUAV show that Alt-UAV achieves the highest success rate among the compared training-free agents across seen and unseen splits. Controlled comparisons further confirm that both the search altitude policy and the combination of candidate selection and inspection contribute to performance.

\bibliographystyle{plainnat}
\bibliography{paper}

\clearpage

\end{document}